\documentclass[10pt,twocolumn,letterpaper]{article}

\usepackage{cvpr} 

\usepackage{times}
\usepackage{epsfig}
\usepackage{graphicx}
\usepackage{amsmath}
\usepackage{amssymb}
\usepackage[pagebackref]{hyperref}

\usepackage{wrapfig}
\usepackage{booktabs}
\usepackage{xcolor}
\usepackage{multicol}
\usepackage{multirow}

\usepackage{svg}
\usepackage{amsmath}
\usepackage{pifont}
\begin{document}

\title{On Model and Data Scaling for Skeleton-based Self-Supervised Gait Recognition}

\author{Adrian Cosma, Andy Cǎtrunǎ, Emilian Rǎdoi\\
University Politehnica of Bucharest\\
Faculty of Automatic Control and Computer Science\\
{\tt\small \{ioan\_adrian.cosma, andy\_eduard.catruna, emilian.radoi\}@upb.ro}
}

\maketitle
\thispagestyle{empty}


\begin{abstract}
   Gait recognition from video streams is a challenging problem in computer vision biometrics due to the subtle differences between gaits and numerous confounding factors. Recent advancements in self-supervised pretraining have led to the development of robust gait recognition models that are invariant to walking covariates. While neural scaling laws have transformed model development in other domains by linking performance to data, model size, and compute, their applicability to gait remains unexplored. In this work, we conduct the first empirical study scaling on skeleton-based self-supervised gait recognition to quantify the effect of data quantity, model size and compute on downstream gait recognition performance. We pretrain multiple variants of GaitPT — a transformer-based architecture — on a dataset of 2.7 million walking sequences collected in the wild. We evaluate zero-shot performance across four benchmark datasets to derive scaling laws for data, model size, and compute. Our findings demonstrate predictable power-law improvements in performance with increased scale and confirm that data and compute scaling significantly influence downstream accuracy. We further isolate architectural contributions by comparing GaitPT with GaitFormer under controlled compute budgets. These results provide practical insights into resource allocation and performance estimation for real-world gait recognition systems.
\end{abstract}
\section{Introduction}
\label{sec:intro}
Modern AI systems scale predictably: more data, more parameters, better performance \cite{kaplan2020scaling,chinchilla2022,caballero2023broken,hernandez2021scaling,muennighoff2023scaling,muscaling}. But does this hold for gait — one of the most subtle and privacy-sensitive biometric modalities? Gait recognition from video streams is a long-standing and difficult problem in the field of computer vision biometrics, due to the subtle differences between gaits across individuals, as well as the innumerable amounts of confounding factors in processing walks \cite{nixon2005gait}. A person's gait is affected, for example, by their clothing, accessories, psychological state and action performed during walking. Furthermore, external factors such as the weather and data acquisition hardware introduce additional measurement errors, as scene lighting, the subject's distance from the camera, video framerate, and viewpoint affect the final representation of the walking sequence. A large body of work has been devoted to explicitly isolate confounding factors, often by the means of constructing specialized architectures \cite{fu2023gpgait,teepe2021gaitgraph,cosma22gaitformer,fvg} or by constructing dedicated training datasets \cite{cosma2024psymo,yu2006framework,Zou2024CCGR,makihara2012ouisir,fvg}. However, for practical deployment of robust gait recognition models, general methods of self-supervised pretraining on an automatically constructed dataset have been developed \cite{cosma2021wildgait,cosma22gaitformer}, which are invariant to walking covariates by sheer exposure to many different walking registers.

\begin{figure}[t]
    \centering
    \includesvg[width=1.0\linewidth]{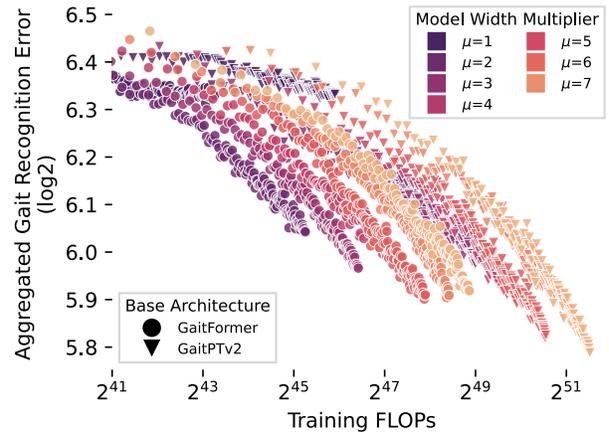}
    \caption{We trained multiple scales of skeleton-based gait recognition models in a self-supervised learning regime on a dataset of 2.7M walking sequences and analyised scaling trends in terms of model size, dataset size and compute allocation.}
    \label{fig:eye-candy}
\end{figure}

More recently, across domains, there has been extensive interest of methods to improve results without the construction of specialized neural architectures, but only through scaling up data and computational resources \cite{cherti2023reproducible,kaplan2020scaling,rosenfeld2019constructive,cherti2023reproducible,caballero2022broken}. Consequently, "the bitter lesson" in statistical representation learning \cite{sutton2019bitter} is that such purely compute-intensive approaches have been vastly outperforming other methods that explicitly integrate human knowledge. These attempts have led to the empirical discovery and mathematical formulation of neural scaling laws \cite{kaplan2020scaling}, in which the relationship between model performance and amount of data or compute is expressed as a power law of the form $L \propto N^{\alpha}$, where $L$ is the model loss or a performance metric such as accuracy, $N$ is the number of parameters or amount of data and $\alpha$ is constant parameters found through curve fitting. Scaling laws are crucial from a practical standpoint as they enable estimation of model performance as a function of compute and data before actual training. This approach leads to reduced costs and better resource allocation, justifying whether potentially costly business decisions of scaling up data or compute are worth the performance improvement. 

To date, a large scale study on the scaling behaviour of self-supervised gait recognition models has not been performed. In this work, we conduct the first comprehensive scaling study of skeleton-based self-supervised gait recognition, quantifying the effects of model size, data volume, and compute allocation on downstream zero-shot performance.  For our study, we collected a dataset of 2.7M skeleton sequences from real-world video streams — the largest such dataset to date — which offers empirical scaling trends grounded in real-world variability. We study scaling properties of skeleton-based gait recognition, as human poses are lightweight, invariant to clothing or walking variations and encode mostly movement patterns as opposed to appearance information \cite{cătrună2024paradox}. We benchmark a modified version of the state-of-the-art skeleton-based gait processing transformer architecture, GaitPT \cite{catruna2023gaitpt}, to utilize the latest improvements in transformer models in terms of scalability and training stability.  Furthermore, we directly compare scaling performance against GaitFormer \cite{cosma22gaitformer}, a strong transformer model, to isolate architectural contributions from raw scale. We analyze compute efficiency by training all models under controlled FLOP budgets and report iso-compute comparisons. We evaluate zero-shot performance on aggregated metrics from 4 different datasets: CASIA-B \cite{yu2006framework}, PsyMo \cite{cosma2024psymo}, GREW \cite{zhu2021gait} and Gait3D \cite{zheng2022gait}, to have a comprehensive view of scaling properties across controlled and in-the-wild gait recognition scenarios.

Our work makes the following contributions:

\begin{enumerate}
    \item We provide \textbf{data scaling laws} for self-supervised gait recognition by pretraining on a dataset of 2.7M gait sequences which we collected from in-the-wild video streams. We show that downstream model performance on controlled and in-the-wild gait recognition benchmarks can be extrapolated from small scale data-bound experiments, supporting the use of power-law scaling in this domain. We also investigate the role of \textbf{data quality} using automatic heuristics to filter low-quality sequences. We show that quality improvements have a measurable positive impact on downstream performance.
    
    \item We derive \textbf{model scaling laws} for self-supervised gait recognition, showing predictable improvements with increasing model size. Our experiments with GaitPTv2 span multiple model scales and dataset sizes. As far as we know, we have the first reproduction of zero-shot hyperparameter transfer, through $\mu$P \cite{yang2022tensor}, in area of self-supervised gait recognition.

    \item We perform a detailed \textbf{compute analysis} of model training, comparing different scaling regimes across FLOP budgets. We show that GaitPTv2 outperforms GaitFormer \cite{cosma22gaitformer} due to its increased use of compute for the same number of parameters.
    
\end{enumerate}

\section{Related Work}
\label{sec:related}
Scaling laws have been observed in statistical learning since Cortes et al. \cite{cortes1993learning} proposed power laws for model performance as a function of data size. However, scaling behaviour has only recently been extensively studied, with the availability of compute and internet-scale datasets. Notably, scaling laws have been extensively explored in large language model training \cite{kaplan2020scaling,chinchilla2022,caballero2023broken,hernandez2021scaling,muennighoff2023scaling,muscaling}, establishing a power law relationship between the number of tokens in the dataset and number of model parameters. Hoffmann et al., \cite{chinchilla2022} formulated scaling laws for compute-optimal training, identifying over-training and under-training regimes for language models, given a compute budget. Furthermore, Hernandez et al. \cite{hernandez2021scaling} formulated scaling laws for transfer learning, in which there is a predictable downstream performance in terms of the ratio between the amount of pretraining data and amount of fine-tuning data. In general, scaling laws fall under two broad categories of \textit{data scaling laws} and \textit{model scaling laws}. Some works \cite{caballero2023broken,mckenzie2023inverse} identified problems in which scaling up does not improve upon downstream performance. Caballero et al. \cite{caballero2023broken} formulated a generalization of scaling laws and showed that a piecewise linear modelling of scaling is more appropriate when analyzing scaling across a wide range of orders of magnitude.

Aside from language modelling, scaling laws have been established for other domains as well, for example, for machine translation \cite{bansal2022data}, masked image modelling \cite{xie2023data}, contrastive language-image learning \cite{cherti2023reproducible} and recommendation models \cite{shin2023scaling}. For computer vision \cite{oquab2023dinov2}, the need for scaling informed further architectural developments in vision transformers \cite{dehghani2023scaling,alabdulmohsin2024getting} for efficient distributed training.
 
\begin{figure*}[ht!]
    \centering
    \resizebox{0.75\linewidth}{!}{
        \includegraphics[width=0.3\linewidth]{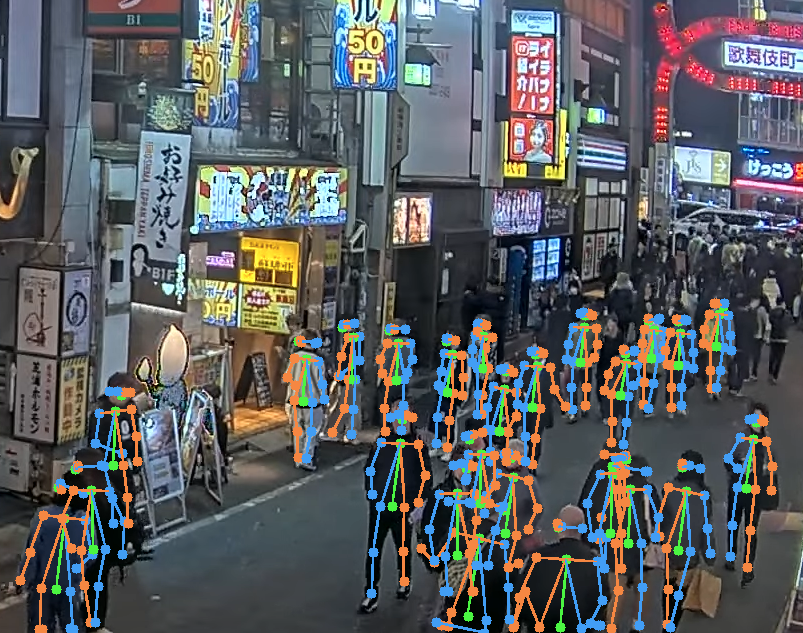}
        \includegraphics[width=0.3\linewidth]{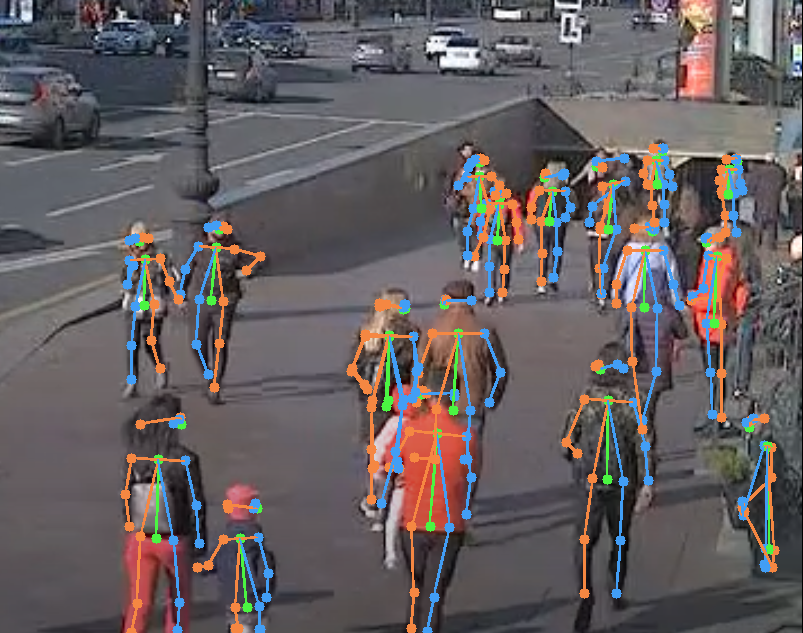}
        \includegraphics[width=0.3\linewidth]{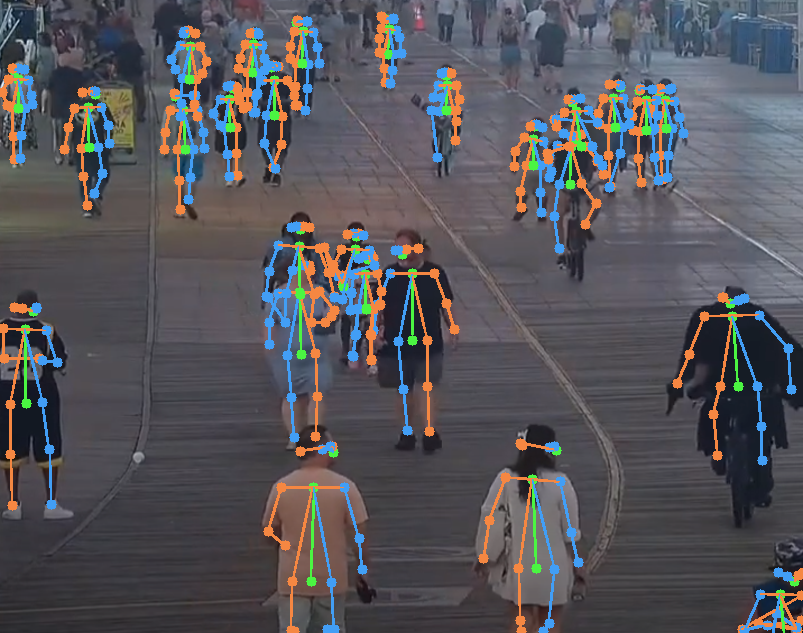}
    }
    \caption{Selected snapshots of different camera feeds used in our dataset annotated with skeleton sequences extracted using a pretrained multi-person pose estimation model. Street webcams in populated areas enable fast and large-scale extraction of gait data.}
    \label{fig:dataset-sample}
\end{figure*}

Regarding gait processing, scaling analysis has not been extensively performed to date. Several large-scale datasets have been proposed \cite{Zou2024CCGR,zhu2021gait,cosma22gaitformer,gaitlu-1m}, but lack the magnitude and diversity for properly studying data scaling in realistic environments. Cosma et al., \cite{cosma2023gaitmorph} combined multiple existing gait datasets into a larger set of 800k sequences for training an autoencoder-type model. However, they did not further explore the effect of the data scale. Previously, works in self-supervised learning have noticed improved downstream performance with increase in data scale \cite{cosma2021wildgait,cosma22gaitformer}, but the amount of data is relatively small to provide insights into scaling behaviour. GaitLU-1M \cite{gaitlu-1m} is a large-scale in-the-wild dataset for gait recognition, but its focus is on modeling gait using sequences of silhouettes, while we are interested in analyzing scaling trends for skeleton-based gait recognition.

{\renewcommand{\arraystretch}{1.0}
    \begin{table*}[ht]
        \small
        \centering
        \caption{Comparison of popular datasets for gait recognition. Our dataset, GREW \cite{zhu2021gait}, Gait3D \cite{zheng2022gait} and DenseGait \cite{cosma22gaitformer}, are collected in the wild and have no clear delimitation between variations and viewpoints. Datasets marked with "$\dagger$" are annotated by their construction in controlled laboratory environments.}
        \resizebox{0.75\linewidth}{!}{
        \begin{tabular}{l|ccccccl}
        \textbf{Dataset} & \textbf{\# IDs} & \textbf{\# Seq.} & \textbf{\# Covariates} & \textbf{\# Views}  & \textbf{Type} & \textbf{Env.} & \textbf{Annotation} \\
        \midrule
        FVG \cite{fvg} & 226 & 2,857 & 5 & 3  & Controlled & outdoor & laboratory$^{\dagger}$ \\
        CASIA-B \cite{yu2006framework} & 124 & 13,640 & 3 & 11 & Controlled & indoor &  laboratory$^{\dagger}$ \\
        PsyMo \cite{cosma2024psymo} & 312 & 14,976 & 7 & 6 & Controlled & indoor & laboratory$^{\dagger}$\\
        OU-ISIR \cite{makihara2012ouisir} & 10,307 & 144,298 & 1 & 14 & Controlled & indoor &  laboratory$^{\dagger}$\\
        CCGR \cite{Zou2024CCGR} & 970 & 1,580,617 & 53 & 33 & Controlled  & indoor &  laboratory$^{\dagger}$ \\
        \midrule
        Gait3D \cite{zheng2022gait} & 4,000 & 25,309 & -- & -- & In the Wild & indoor & manually labeled \\
        GREW \cite{zhu2021gait} & 26,000 & 128,000 & -- & --  & In the Wild & outdoor &  manually labeled \\
        UWG \cite{cosma2021wildgait} & 38,502 & 38,502 & -- & -- & In the Wild & outdoor & unlabeled \\
        DenseGait \cite{cosma22gaitformer} & 217,954 & 217,954 & -- & -- & In the Wild & outdoor &  unlabeled\\
        GaitLU-1M \cite{gaitlu-1m} & 1,035,309 & 1,035,309 & -- & -- & In the Wild & outdoor &  unlabeled\\
        \midrule
        \textbf{Ours} & \textbf{2,779,774} & \textbf{2,779,774} & -- & -- & In the Wild & outdoor & unlabeled \\
        \end{tabular}
        }
        \label{tab:dataset-comparison}
    \end{table*}
}

In this work, we provide an empirical analysis of scaling behaviour for both data size, parameter count and compute utilization for gait recognition in the regime of self-supervised contrastive learning, on a large-scale in-the-wild dataset of 2.7M walking sequences.


\section{Experimental Setup}
\label{sec:method}
In this section we detail our self-supervised pretraining setup, which we describe the collection of a large unlabeled gait recognition dataset, training configuration, the choice of transformer architectures used for gait processing, model evaluation and modeling power-law relationship between downstream performance and scaling parameters.

\subsection{Collection of the Pretraining Dataset}

Since one of our goals is to estimate the effect of scaling up unlabeled gait data to the zero-shot performance of gait recognition models across different model scales, we gather a substantial gait recognition dataset used for self-supervised pretraining. Manually annotating gait sequences requires extensive labor, and is unfeasible for scaling up data to multiple orders of magnitude. Consequently, we aim to learn informative walking representations without manual annotations.

In a manner similar to other approaches \cite{cosma2021wildgait,cosma22gaitformer}, we process publicly available outdoor video streams, each containing a considerable amount of people walking, as exemplified in Figure \ref{fig:dataset-sample}. The videos are chosen to have a diverse range of weather conditions, times of day, camera viewpoints, geographic locations, and containing both static and moving cameras.  Works in gait recognition employ either sequences of silhouettes \cite{chao2019gaitset,lin2022gaitgl}, body meshes \cite{li2020end,zheng2022gait} or sequences of skeletons to encode walking \cite{teepe2021gaitgraph,catruna2023gaitpt,cosma22gaitformer}. We chose to use skeletons as they are easy to accurately extract \cite{alphapose,xu2022vitpose}, lightweight in terms of storage and processing, and methods using skeleton sequences have shown promising results in this area \cite{cosma22gaitformer,catruna2023gaitpt,fu2023gpgait,fan2024skeletongait}. Furthermore, skeletons enable fine-grained control on data quality by offering information for each joint across time. While there are methods that also incorporate skeleton maps \cite{fan2024skeletongait} and SMPL body meshes \cite{zheng2022gait}, we opted for the simplest case in which only sequences of skeletons are used to compute gait representations.

We process the stream to extract human poses using AlphaPose \cite{alphapose} and track each pose in the video using SortOH \cite{nasseri2021simple}.  We employed minimal filtering of the extracted skeleton sequences, by keeping only sequences of a minimum of 48 frames (at a framerate of 24fps), above an average joint confidence threshold of 0.6.  

Our dataset contains a diverse range of walking registers; for instance, pedestrians are walking wearing different pieces of clothing or footwear, walking while carrying luggage or shopping bags, walking alongside other people, walking while talking on the phone or doing other actions. Each person's identity is anonymized: we discard any appearance cues and metadata and keep only the skeleton sequence. In total, our dataset contains 2.7M skeleton sequences, with an average walking duration of 168 frames for a total of 132,931 days of walking. It is currently the largest in-the-wild and unlabeled dataset reported in literature, having an order of magnitude more skeleton sequences compared to previous in-the-wild datasets \cite{zhu2021gait,cosma22gaitformer}. Table \ref{tab:dataset-comparison} shows a comparison with other gait recognition datasets from literature. Our dataset is private and we only use it for unsupervised pretraining to gauge the effect of scaling data on downstream zero-shot performance.

\subsection{Contrastive Self-Supervised Gait Recognition}

For self-supervision, a natural pretraining regime for this domain is contrastive learning, in which the model learns to separate walking sequences of different people and aggregate walks of the same person. This approach to gait recognition has been done in the past for both label and unlabeled datasets \cite{catruna2023gaitpt,cosma22gaitformer}. In particular, we adopt the SimCLR \cite{chen2020simple} approach for contrastive learning, in which we augment a walking sequence in two different ways to form positive pairs. Following previous works \cite{cosma22gaitformer,catruna2023gaitpt}, we used the following augmentations for skeleton sequences: random temporal crops, random flips, random mirror, joint noise, random paces and randomly smoothing the sequence. We translate and scale each sequence based on the skeleton in the middle of the sequence, adopting the "sequence normalization" approach formulated by Catruna et al. \cite{cătrună2024paradox}, since it has minimal impact upon in-the-wild gait recognition scenarios. Compared to other works \cite{fu2023gpgait}, we do not use any explicit anthropomorphic features (e.g., limb lengths, limb angles) that may provide shortcuts and artificially increase recognition performance in certain benchmarks \cite{cătrună2024paradox}.

Additionally, alongside the contrastive loss, we use the KoLeo regularizer \cite{sablayrolles2018spreading} to induce uniform feature spreading within a batch. The KoLeo regularizer improves performance in retrieval-type tasks \cite{sablayrolles2018spreading}, and has been used in self-supervised learning on images \cite{oquab2023dinov2,caron2020unsupervised}. Given a batch of $k$ feature vectors $(f_1, f_2, \dots f_k)$, it is defined as $\mathcal{L}_{koleo} = - \frac{1}{n} \sum^n_{i = 1}\log(d_{n,i})$, where $d_{n,i} = \min\limits_{j \neq i}\lVert x_j - x_i \rVert$, the minimum distance between $x_i$ and each of the other vectors within the batch. As such, the pretraining loss in our setting is defined as $\mathcal{L} = \mathcal{L}_{\text{SimCLR}} + \lambda \mathcal{L}_{koleo}$, where we chose $\lambda = 0.01$. 

\subsection{Transformer Architectures for Gait Processing}

\begin{table}[hbt!]
    \centering
    \caption{Architectural details of the deep and thin GaitPTv2. We show size configurations for both the spatial and temporal transformer layers at each of the four GaitPT stages. GFLOPs are computed for a forward pass with batch size of 1.}
    \resizebox{1.0\linewidth}{!}{
        \begin{tabular}{cc|crrc|rr}
            & \textbf{Model Name} & \textbf{Depth} & \textbf{d$_{model}$} & \textbf{n\_heads} & \textbf{Output Emb.} & \textbf{\# Params} & \textbf{GFLOPs} \\
            \midrule
         \multirow{7}{*}{\rotatebox{90}{\textbf{Deep \& Thin}}} & GaitPTv2-1 & \{2, 2, 12, 2\} & \{4, 8, 16, 32\} & \{1, 2, 4, 8\} & 32 & 154,696 & 0.036 \\
         & GaitPTv2-2 & \{2, 2, 12, 2\} & \{8, 16, 32, 64\} & \{2, 4, 8, 16\} & 64 & 607,484 & 0.145\\
         & GaitPTv2-3 & \{2, 2, 12, 2\} & \{16, 32, 64, 128\} & \{4, 8, 16, 32\} & 128 & 2,408,356 & 0.582\\
         & GaitPTv2-4 & \{2, 2, 12, 2\} & \{24, 48, 96, 192\} & \{6, 12, 24, 48\} & 192 & 5,402,956 & 1.308 \\
         & GaitPTv2-5 & \{2, 2, 12, 2\} & \{32, 64, 128, 256\} & \{8, 16, 32, 64\} & 256 & 9,591,284 & 2.325\\
         & GaitPTv2-6 & \{2, 2, 12, 2\} & \{48, 96, 192, 384\} & \{12, 24, 48, 96\} & 384 & 21,549,124 & 5.229 \\
         & GaitPTv2-7 & \{2, 2, 12, 2\} & \{64, 128, 256, 512\} & \{16, 32, 64, 128\} & 512 & 38,281,876 & 9.295 \\
         \midrule
        \multirow{7}{*}{\rotatebox{90}{\textbf{Shallow \& Wide}}} & GaitPTv2-1 & \{2, 2, 4, 1\} & \{16, 64, 64, 128\} & \{1, 4, 4, 8\} & 64 & 1,194,144 & 0.246 \\
        & GaitPTv2-2 & \{2, 2, 4, 1\} & \{32, 128, 128, 256\} & \{2, 8, 8, 16\} & 128 & 4,754,208 & 0.982 \\
        & GaitPTv2-3 & \{2, 2, 4, 1\} & \{48, 192, 192, 384\} & \{3, 12, 12, 24\} & 192 & 10,680,736 & 2.210 \\
        & GaitPTv2-4 & \{2, 2, 4, 1\} & \{64, 256, 256, 512\} & \{4, 16, 16, 32\} & 256 & 18,973,728 & 3.929 \\
        & GaitPTv2-5 & \{2, 2, 4, 1\} & \{80, 320, 320, 640\} & \{5, 20, 20, 40\} & 320 & 29,633,184 & 6.138 \\
        & GaitPTv2-6 & \{2, 2, 4, 1\} & \{96, 384, 384, 768\} & \{6, 24, 24, 48\} & 384 & 42,659,104 & 8.839 \\
        & GaitPTv2-7 & \{2, 2, 4, 1\} & \{112, 448, 448, 896\} & \{7, 28, 28, 56\} & 448 & 58,051,488 & 12.03 \\
        \midrule
        \multirow{7}{*}{\rotatebox{90}{\textbf{Non-Hierarchical}}} & GaitFormer-1 & 9 & 64 & 4 & 64 & 605,008 & 0.039 \\
        & GaitFormer-2 & 9 & 128 & 8 & 128 & 2,402,688 & 0.156 \\
        & GaitFormer-3 & 9 & 192 & 12 & 192 & 5,393,328 & 0.351 \\
        & GaitFormer-4 & 9 & 256 & 16 & 256 & 9,576,928 & 0.624 \\
        & GaitFormer-5 & 9 & 320 & 20 & 320 & 14,953,488 & 0.974 \\
        & GaitFormer-6 & 9 & 384 & 24 & 384 & 21,523,008 & 1.403 \\
        & GaitFormer-7 & 9 & 448 & 28 & 448 & 29,285,488 & 1.909 \\
        \end{tabular}
    }
    \label{tab:gaitpt-sizes}
\end{table}

In our experiments, we used slightly modified variants of GaitPT \cite{catruna2023gaitpt} and GaitFormer \cite{cosma22gaitformer}. GaitPT \cite{catruna2023gaitpt} is a hierarchical skeleton transformer with good results for skeleton-based gait recognition on multiple benchmarks. Multiple works in gait recognition \cite{catruna2023gaitpt,fu2023gpgait} have recognized the need of hierarchical processing of skeleton sequences for achieving good performance, by aggregating low-level coordinate information to high level limb movements. We chose GaitPT as a representative architecture for a larger class of models hierarchical pose-based gait models \cite{fu2023gpgait}. In contrast to other architectures \cite{liao2020model,fu2023gpgait}, GaitPT is a fully attention-based model \cite{vaswani2017attention}, which benefits from known scaling properties \cite{yang2022tensor,kaplan2020scaling,dehghani2023scaling} and parallelization of transformer models. In particular, GaitPT is organized similarly to SwinTransformers \cite{liu2021swin}, having 4 sequential stages, each stage having spatial transformer layers operating on spatial dimensions of each skeleton, and a temporal transformer layers aggregating temporal information of the sequence. Readers are referred to the work of Catruna et al. \cite{catruna2023gaitpt} for a more detailed description of the model. GaitFormer \cite{cosma22gaitformer} is another full-attention architecture, inspired by vision transformers \cite{dosovitskiy2021an}, in which a simple transformer encoder is used to process sequences of skeletons to output gait representations. In this case, there is no hierarchy of representations, and each skeleton is treated as a single token.

We modify the original GaitPT and GaitFormer implementations to adopt several transformer improvements \cite{rms-norm,gpt-j,shazeer2020glu,dehghani2023scaling,su2024roformer} for training stabilization and higher throughput without loss of expressive power. 
In particular, we used "parallel layers" \cite{gpt-j} by applying the Attention and MLP blocks in parallel, instead of sequentially as in the standard Transformer \cite{vaswani2017attention}, we removed bias of QKV projection layers \cite{dehghani2023scaling}, changed LayerNorm layers to RMS normalization \cite{rms-norm}, changed the activation function from GeLU \cite{hendrycks2016gaussian} to SwiGLU \cite{shazeer2020glu} and we used Rotary Positional Embeddings \cite{su2024roformer} instead of absolute positional embeddings. For GaitPT, the rotary positional embeddings are instantiated per stage. Furthermore, in the original GaitPT implementation, the final embedding is the direct concatenation of outputs from all 4 stages, resulting in a very large dimensionality -- here, we project with a linear layer each stage ouput to a vector of dimension $emb\_size$. As such, the output concatenation of each stage has dimension $4\cdot emb\_size$, which is projected using another linear layer into dimension $emb\_size$. We name this modified model GaitPTv2. Following the training procedure from SimCLR \cite{chen2020simple}, we used a non-linear output projection head during training for both architectures.

\noindent \textbf{Model Configurations} Since we are interested in analyzing the impact of different model sizes (measured by number of trainable parameters), we vary only the width, and keep the depth fixed, allowing the use of maximal update parametrization ($\mu$P) \cite{yang2021tuning} for zero-shot hyperparameter transfer across model widths. 

In this work, we analyze two variants of the GaitPTv2 architecture: a deeper but thinner model and a more shallow but wider model (Table \ref{tab:gaitpt-sizes}). Since the original GaitPT is comprised of 4 stages, we vary the GaitPTv2 model size by fixing its depth across each stage and increasing only the transformer model widths and number of heads in terms of a single multiplicative factor $c$: $d_{model}^{(c)} = c \cdot d_{model}^{base}$ and n\_heads$^{(c)} = c \cdot n\_heads^{base}$. For GaitFormer, we build each model configuration using the corresponding hyperparameters from the third GaitPT stage, as shown in Table \ref{tab:gaitpt-sizes}. For the deeper GaitPTv2, we used 12 layers in Stage 3 and 2 layers everywhere else, inline with SwinTransformers \cite{liu2021swin}. For the shallower model, we used only 4 layers in Stage 3, but increased d$_{model}$ for all stages. 

\subsection{Hyperparameters and Pretraining Details}
\label{sec:pretraining}

We used $\mu$P parametrization \cite{yang2021tuning}, to avoid expensive hyperparameter search in larger models. $\mu$P parametrization modifies the learning rate and initializations for feed-forward layers in proportion to the relative width compared to a base model. As such, the optimal learning rate found for a base model size can be directly adapted to larger model scales, as long as the depth of the model remains fixed. In this way, our trained models are not affected by sub-optimal hyperparameter choices and enable us to make a fair representation of model performance across scales. Other works in this area \cite{xie2023data} fix hyperparameters for all model scales, which is sub-optimal, since it considerably affects results. As far as we know, this work is the first reproduction of $\mu$P in self-supervised gait recognition.

As opposed to scaling analyses in NLP \cite{kaplan2020scaling}, we do not limit model training on a single epoch, since transformers usually benefit from processing multiple epochs \cite{muennighoff2023scaling,xie2023data}. Furthermore, contrastive learning pretraining schemes such as SimCLR \cite{chen2020simple} require multiple training epochs to achieve a reasonable performance due to the increased diversity of data augmentations. 

For modeling power-law scaling of model and data size, we train our models for a fixed number of 25 training epochs across data subsets and model scales. We used a fixed batch size of 256 samples across all model scales, which includes the 2 augmented views for each walking sequence. For modeling compute allocation, we increased the batch size to 2048 for all models. All models are trained using AdamW optimizer \cite{kigma2014adam} with mixed-precision, and we used a learning rate of 0.0016 for the smallest model (which is adapted using $\mu$P across model scales) with a cosine learning rate schedule with 1024 iterations for warm-up. We used two NVIDIA H100 / A100 GPUs for training.

\subsection{Model Evaluation: Controlled and In-the-Wild}

For computing model performance across scales, we evaluated the pretrained model in a zero-shot manner (with no fine-tuning) on 4 datasets: CASIA-B \cite{yu2006framework}, PsyMo \cite{cosma2024psymo}, GREW \cite{zhu2021gait} and Gait3D \cite{zheng2022gait}. For performance evaluation in controlled gait recognition settings we used CASIA-B and PsyMo, and for performance evaluation in realistic scenarios (i.e., "in-the-wild") we used GREW and Gait3D. Both CASIA-B and PsyMo have similar dedicated evaluation procedures that aim to have a fine-grained measure of performance across viewpoints and walking variations. Readers are referred to each dataset's paper for a detailed description of the evaluation protocol. For controlled scenarios, we follow each dataset's evaluation protocol and compute the average performance across viewpoints and scales, excluding identical view cases, and aggregate both metrics in a single value. For in-the-wild performance evaluation, for GREW and Gait3D, we follow each dataset's evaluation procedure and compute rank-5 recognition accuracy and aggregate both metrics in a single value.

\subsection{Modeling the Power-law for Gait Recognition}

We hypothesize that gait recognition follows power-law scaling similar to language modeling \cite{kaplan2020scaling,chinchilla2022} and other domains \cite{xie2023data,cherti2023reproducible}. In the case of contrastive learning for gait recognition, we chose to model performance in terms of aggregated downstream accuracy instead of relying on loss value. Consequently, we assume that the aggregated performance $P(N, D)$ in terms of accuracy of a gait recognition model depends on the number of model parameters $N$ and the dataset size $D$ measured in number of skeleton sequences. Following similar works \cite{kaplan2020scaling}, we assume that $P(N, D) = N^{\alpha} + D^{\beta} + E$, where $E$ is a constant irreducible error term. In our experiments, we compute separate scaling laws for model and dataset scale, respectively. If we fix the model size into a fixed set of scales and varied the dataset size, the $N^{\alpha}$ term becomes constant, and the data scaling law becomes $P_{N}(D) = D^{\beta} + E$. Similarly, if we fixed the dataset size, the model scaling law becomes $P_{D}(N) = N^{\alpha} + E$. The parameters $\alpha$ and $\beta$ can be found through least-squares linear regression on a set of model performance values across scales, in a log-log plot.


\section{Results}
\subsection{Power-Law Scaling of GaitPTv2}
\label{sec:results-powerlaws}

\begin{figure}[hbt!]
    \centering
    \includesvg[width=0.8\linewidth]{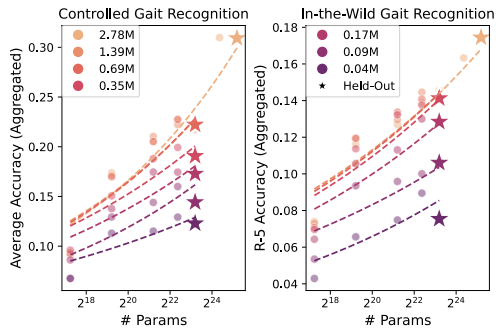}
    \caption{Scaling trends for increasing the model size by parameter count, across multiple dataset sizes. We compute scaling trends only on the data points marked with a "$\bullet$" symbol, while the "$\star$" data point is used for validation. Increasing the parameter count yields a predictable  positive increase in performance.}
    \label{fig:model-trend}
\end{figure}

In this subsection, we present power-law scaling trends and analyze extrapolations across model and data scales. Here, we used the \textit{\textbf{"Deep \& Thin"}} variant of GaitPTv2.

\noindent \textbf{Model scaling for self-supervised gait recognition.} In Figure \ref{fig:model-trend} we show model scaling behaviour for several sizes of the pretraining dataset. Larger models have almost always better performance, regardless of the amount of pretraining data. When computing trend lines, we used all but the last data point, and used the final training run (denoted by the $\star$) for validation. We only train GaitPTv2-7 on the largest data subset, due to the computational constraints of our setup. The largest model's performance (i.e., $\mu = 7$ and $\mu = 6$) closely follows the trend line and their performance can be extrapolated from training smaller scale models. 

\begin{table}[hbt!]
    \centering
    \caption{Scaling parameters $\alpha$ and $\beta$ for both model size and dataset size in zero-shot controlled and in-the-wild gait recognition scenarios.}
    \resizebox{1.0\linewidth}{!}{
    \begin{tabular}{lcc||lcc}
    \multicolumn{3}{c}{\textbf{Model Scaling}} & \multicolumn{3}{c}{\textbf{Data Scaling}} \\
    \textbf{Dataset Size (\# Sequences)} & \textbf{$\alpha$ (Controlled)} & \textbf{$\alpha$ (In-the-Wild)} & \textbf{\# Parameters} & \textbf{$\beta$ (Controlled)} & \textbf{$\beta$ (In-the-Wild)} \\
    \midrule
    43.4K (1.56\%) & 0.100 & \textbf{0.118} & 0.15M ($\mu = 1$) & 0.070 & 0.094 \\
    86.9K (3.12\%) & 0.139 & 0.099 & 0.61M ($\mu = 2$) & 0.078 & 0.100 \\
    173.7K (6.25\%) & 0.122 & 0.111 & 2.41M ($\mu = 3$) & \underline{0.109} & \underline{0.102} \\
    347.5K (12.5\%) & 0.124 & \underline{0.113} & 5.4M ($\mu = 4$) & 0.102 & 0.091 \\
    694.9K (25.0\%) & \underline{0.148} & \textbf{0.113} & 9.59M ($\mu = 5$) & \textbf{0.114} & \textbf{0.108} \\
    1389.9K (50.0\%) & 0.145 & 0.109 & \\
    2779.8K (100.0\%) &\textbf{0.164} & 0.112 & \\
    \end{tabular}
    }
    \label{tab:trends}
\end{table}

In Table \ref{tab:trends} on the left-hand side, we show model scaling parameters $\alpha$ for multiple subsets of our dataset. Even though training on the largest subset has the steepest scaling parameter for controlled scenarios, for in-the-wild settings parameters are fairly close to one another. This is due to the fact that the models are likely undertrained and could significantly benefit from more training. Training on smaller amounts of data usually results in overfitting, and by observing good results for smaller models in this scenario indicates that the models are not trained to saturation. 

\begin{figure}[ht!]
    \centering
    \includesvg[width=0.8\linewidth]{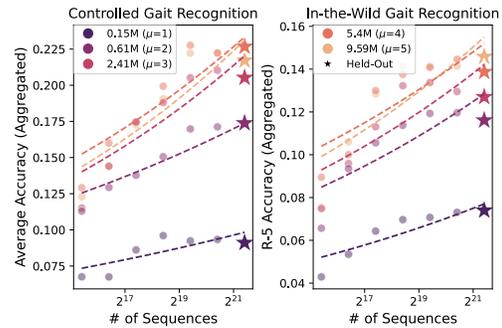}
    \caption{Scaling trends for increasing the dataset size in terms of the number of skeleton sequences, across multiple model sizes. We compute scaling curves from points marked "$\bullet$". The point marked with "$\star$" is used for validation. Increasing the dataset size yields a predictable positive increase in performance.}
    \label{fig:data-trend}
\end{figure}

\noindent \textbf{Data scaling for self-supervised gait recognition.} In Figure \ref{fig:data-trend} we show data scaling behaviour for several model sizes when trained with progressively larger dataset sizes. All model scales benefit from increasing the size of the pretraining dataset in both controlled and in the wild scenarios. When computing the trend line, we used all but the largest data scale, and the final training run (denoted by the $\star$) for validation. While the final point is close to the trend line, there seems to be a saturation point at larger amounts of data, likely due to the added noise in the dataset. We explore the effect of data quality on scaling below. We could argue that gait has comparatively less entropy than natural language, where scaling has been more extensively studied \cite{kaplan2020scaling}, which might lead to faster data saturation \cite{caballero2022broken}. As opposed to collecting text data \cite{villalobos2022will}, diverse gait data is easier to gather from video streams, as different environments may lead to radically different ways of walking. In Table \ref{tab:trends}, on the right-hand side, we show the data scaling parameters $\beta$ for multiple model sizes. From our experiments, the larger the model size, the more it is able to consume data for this task. The model is likely under-trained and could significantly improve its performance with training to saturation.

\begin{figure}[hbt!]
    \centering
    \includesvg[width=0.8\linewidth]{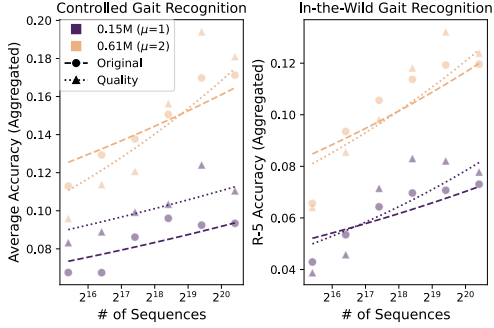}
    \caption{Comparison of data scaling behaviour between models trained on high quality samples versus models trained on samples from the original set.}
    \label{fig:quality-trend}
\end{figure}

\noindent \textbf{Effect of skeleton data quality} 
To study the effect of modifying the data quality on scaling trends, we propose a simple heuristic to order skeleton sequences in terms of the quality of extracted human poses. Considering that a 2D skeleton sequence $S$ is comprised of a set of $J = 18$ joints having 2 coordinate values $(x_j, y_j)$ and a confidence score $c_j$, we compute, for each joint, the average confidence $\overline{c_j}$ and the variance of the confidence $\sigma_{c_j}$ across sequence length. The quality score is defined as $Q_S = \sum_{j=0}^J (\overline{c_j} - \log{\sigma_{c_j}})$. Assuming high quality sequences should have high average confidence and low confidence variance across time, ordering the dataset by $Q_S$ gives a monotonically increasing set of skeleton sequences by quality of extraction. As such, in each subset we sample the top quality sequences. Figure \ref{fig:quality-trend} shows the data scaling properties of training with high quality samples compared to the original, randomly sampled subset. Training with high quality samples yields consistent better performance and a slightly steeper scaling curve. For example, for $\mu = 2$, the scaling parameter $\beta$ for the high quality samples is 0.1319 versus 0.078 on controlled scenarios, and 0.125 versus 0.099 in in-the-wild scenarios. 


\subsection{Scaling the compute budget across architectures}
\label{sec:results-flops}
In this subsection, we present our analyses in terms of amount of compute (FLOPs), and provide a comparison between GaitPTv2 and GaitFormer. Here, we used the \textit{\textbf{"Shallow \& Wide"}} variant of GaitPTv2.



\begin{figure}[hbt!]
    \centering
    \includesvg[width=0.8\linewidth]{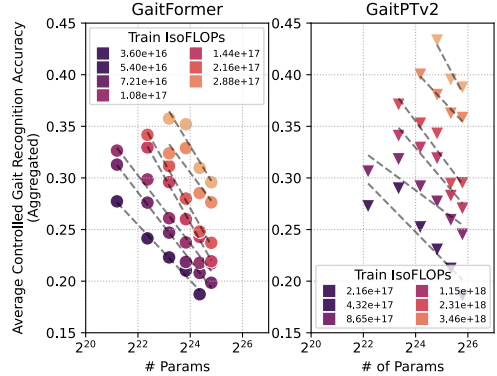}
    \caption{Training IsoFLOPs curves for GaitPTv2 and GaitFormer, comparing number of parameters and controlled gait recognition accuracy. The points on each curve utilize approximately the same amount of training compute. Best viewed in color.}
    \label{fig:params-isoflops}
\end{figure}

\begin{figure}[hbt!]
    \centering
    \includesvg[width=0.8\linewidth]{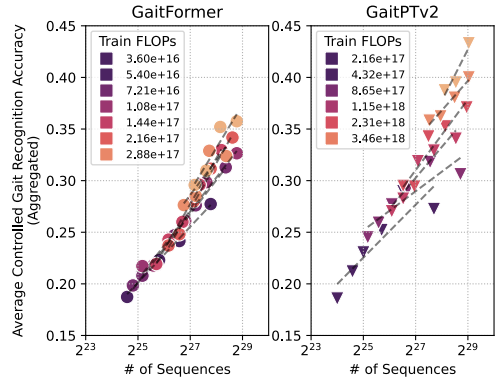}
    \caption{Training IsoFLOPs curves for GaitPTv2 and GaitFormer, comparing number of training gait sequences and controlled gait recognition accuracy. The points on each curve utilize approximately the same amount of training compute. Best viewed in color.}
    \label{fig:data-isoflops}
\end{figure}

In Figure \ref{fig:params-isoflops} we show training IsoFLOPs curves for GaitPTv2 and GaitFormer. We fixed compute budgets and selected the closest model that reached that budget during training. Each curve was obtained by fitting a linear model to model accuracy as function of number of parameters: $P(\alpha) = \alpha\log_2(N) + E$, where $P$ is downstream controlled gait recognition accuracy, $N$ is the number of parameters and $E$ is the intercept. In our scenario, \textit{smaller models have a more efficient use of compute given enough data}. Similarly, in Figure \ref{fig:data-isoflops}, we plot training IsoFLOPs curves obtained by fitting a linear model in the form $P(\beta) = \beta\log_2(D) + E$, where $D$ is the number of training gait sequences. In this case, \textit{increasing the number of gait sequences seen during training improves performance across compute budgets}.

\begin{figure}[hbt!]
    \centering
    \includesvg[width=0.9\linewidth]{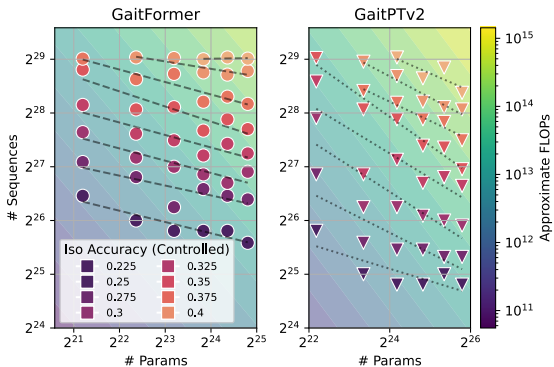}
    \caption{Iso-Accuracy curves on controlled gait recognition for GaitPTv2 and GaitFormer. The background is colored using IsoFLOPs contours for both models. Best viewed in color.}
    \label{fig:iso-accuracy}
\end{figure}

\begin{figure}[hbt!]
    \centering
    \includesvg[width=\linewidth]{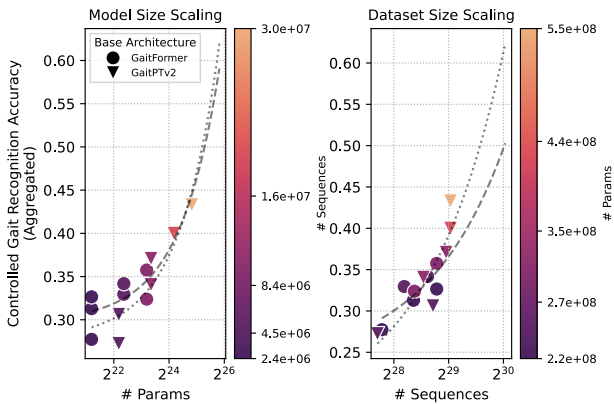}
    \caption{Trends for using the most efficient models across FLOP budgets. Increasing dataset size better differentiates between the scaling dynamics of GaitPTv2 and GaitFormer.}
    \label{fig:params-compute-optimal}
\end{figure}

In Figure \ref{fig:iso-accuracy} we show Iso-Accuracy curves for GaitPTv2 and Gaitformer as well as Iso-FLOP contours. We approximate the amount of FLOPs consumed by each model as $6.5 \cdot ND$ for GaitPTv2 and $\sim 2 \cdot ND$ for GaitFormer, by fitting a linear model of the form $C(\gamma) = \gamma ND$. Based on these results, GaitFormer, a simple transformer encoder model, is more efficient in terms of compute for the same number of parameters compared to GaitPTv2. However, \textit{GaitPTv2 obtains better accuracy because it uses more compute.} The main difference stems from the way skeleton sequences are processed between the two models. In the case of GaitFormer, each "token" is considered a flattened skeleton, which discards explicit spatial relationship between joints. GaitPTv2, however, processes sequences hierarchically \cite{catruna2023gaitpt,liu2021swin}, from single joints to body parts, having a larger effective context length, and, in turn, more compute expenditure per sequence. As a consequence, \textit{GaitPTv2 scales better with amount of data compared to GaitFormer}. We show this result in Figure \ref{fig:params-compute-optimal}: we select the most efficient models in terms of compute (from Figures \ref{fig:params-isoflops} and \ref{fig:data-isoflops}) and plot trends across parameter counts and number of training sequences. Scaling parameter count does not show a substantial difference between models, but the gap is evident when scaling number of training gait sequences: GaitPTv2 is using more compute per gait sequence, resulting in better downstream gait recognition performance.

\begin{figure}[hbt!]
    \centering
    \includesvg[width=0.7\linewidth]{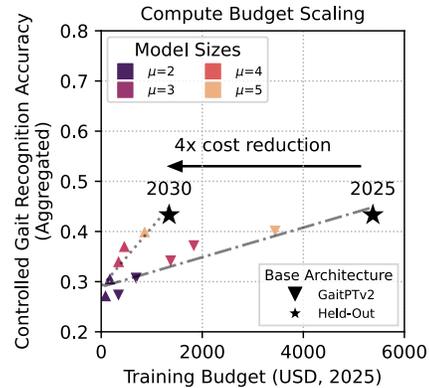}
    \caption{Scaling trends for increasing the dollar cost of training. We compute scaling curves from points marked "$\blacktriangledown$ / "$\blacktriangle$". The point marked with "$\star$" is used for validation. Incorporating Moore's Law yields a sharper scaling parameter across time.}
    \label{fig:dollars-2025}
\end{figure}

\noindent \textbf{Dollar Cost of Training} For practical applications, it is paramount to estimate the accuracy of a training run for a given monetary budget before actually training, since training large models can incur a significant cost.  Considering the cost of a GFLOP to be $\$0.03$ in 2017 \footnote{\href{https://humanprogress.org/trends/vastly-cheaper-computation/}{https://humanprogress.org/trends/vastly-cheaper-computation/}, Accessed: 11 April 2025}, we show in Figure \ref{fig:dollars-2025} the dollar cost of training self-supervised gait recognition models and extrapolate the accuracy trends for a given held-out budget. The fitted line can accurately estimate the downstream accuracy. Further, Moore's law \cite{moores-law} should be taken into account when estimating future costs in terms of compute. Moore's law can be expressed as $C(t) = C_0 \cdot 2^{-\frac{t}{T}}$, where $T$ is the halving period of cost (here, $T = 2.5$) and $C_0 = \$0.03$. Incorporating this trend we obtain a 4x reduction in dollar cost for training in 5 years. In other words, \textit{simply waiting 5 years will increase accuracy by around 10\% for a fixed cost budget.}

\section{Conclusions}
\label{sec:conc}
Our study represents the first attempt at constructing scaling laws for self-supervised gait recognition, exploring the dynamics of model performance, data quantity, and computational resources to downstream zero-shot performance on both controlled gait recognition scenarios \cite{yu2006framework,cosma2024psymo} and in-the-wild scenarios \cite{zheng2022gait,zhu2021gait}. We gathered a dataset of 2.7M skeleton sequences in-the-wild, the largest reported in literature, and used an improved version of GaitPT \cite{catruna2023gaitpt} for pretraining. We showed that, power-laws performance trends do apply to gait recognition, enabling practitioners to predict the performance of models by training only smaller versions of models, and on smaller amounts of data. We also presented the first reproduction of $\mu$P \cite{muscaling} in gait recognition, which allowed us to directly transfer hyperparameters from small models to larger ones without additional search. We further isolate architectural contributions by comparing GaitPTv2 with GaitFormer under controlled compute budgets and showed that GaitPTv2 scales better with data because, by its hierarchical design, it is able to use more compute per gait sequence. Our work represents a promising avenue for further research into scaling self-supervised gait recognition.

{\small
\bibliographystyle{ieee}
\bibliography{refs}
}

\end{document}